\documentclass[sigplan]{acmart}[9pt]
\AtBeginDocument{%
  \providecommand\BibTeX{{%
    \normalfont B\kern-0.5em{\scshape i\kern-0.25em b}\kern-0.8em\TeX}}}

\usepackage{url}
\usepackage{enumitem}

\usepackage[utf8]{inputenc} % allow utf-8 input
\usepackage[T1]{fontenc}    % use 8-bit T1 fonts
\usepackage{hyperref}       % hyperlinks
\usepackage{url}            % simple URL typesetting
\usepackage{booktabs}       % professional-quality tables
\usepackage{amsfonts}       % blackboard math symbols
\usepackage{nicefrac}       % compact symbols for 1/2, etc.
\usepackage{microtype}      % microtypography
\usepackage{xcolor}         % colors
\usepackage{colortbl}
\usepackage{subcaption}
\usepackage{graphicx}
\usepackage{svg}
\usepackage{multirow}
\usepackage{wrapfig}
\usepackage{threeparttable}
\usepackage{tikz}
\usepackage{makecell}
\usepackage{tablefootnote}

\usepackage{color}% this one is added by hang
\begin{document}

%% \vspace{-12pt}
\title{High-Level Synthesis Performance Prediction using GNNs: Benchmarking, Modeling, and Advancing}

\settopmatter{printacmref=false} % Removes citation information below abstract
\renewcommand\footnotetextcopyrightpermission[1]{} % removes footnote with conference information in first column
\pagestyle{plain} % removes running headers
%\affiliation{%
%  \institution{Institute for Clarity in Documentation}
%  \streetaddress{P.O. Box 1212}
%  \country{USA}

%\iffalse
\author{Nan Wu}
\affiliation{%
\small
nanwu@ucsb.edu \\
  \institution{University of California, Santa Barbara}
  \city{Santa Barbara}
  \state{CA}
  \country{USA}
  \postcode{93106}
}
\author{Hang Yang}
\affiliation{%
\small
innallyyang@hotmail.com \\
  \institution{Nankai University}
  \city{Tianjin}
  %\state{CA}
  \country{China}
  %\postcode{93106}
}
\author{Yuan Xie}
\affiliation{%
\small
  yuanxie@ucsb.edu \\
  \institution{University of California, Santa Barbara}
  \city{Santa Barbara}
  \state{CA}
  \country{USA}
  \postcode{93106}
}
\author{Pan Li}
\affiliation{%
\small
panli@purdue.edu \\
  \institution{Purdue University}
  \city{West Lafayette}
  \state{IN}
  \country{USA}
  \postcode{47907}
}
\author{Cong Hao}
\affiliation{%
\small
callie.hao@ece.gatech.edu \\
  \institution{Georgia Institute of Technology}
  \city{Atlanta}
  \state{GA}
  \country{USA}
  \postcode{30332}
}
% \fi

\begin{abstract}
Agile hardware development requires fast and accurate circuit quality evaluation from early design stages.
Existing work of high-level synthesis (HLS) performance prediction usually needs extensive feature engineering after the synthesis process.
To expedite circuit evaluation \textit{from as earlier design stage as possible}, we propose a rapid and accurate performance modeling, exploiting the representation power of graph neural networks (GNNs) by representing C/C++ programs as graphs.
The contribution of this work is three-fold.
\underline{First}, we build a standard benchmark containing 40k C synthesizable programs, which includes both synthetic programs and three sets of real-world HLS benchmarks. Each program is implemented on FPGA to generate ground-truth performance metrics.
\underline{Second}, we formally formulate the HLS performance prediction problem on graphs, and propose multiple modeling strategies with GNNs that leverage different trade-offs between prediction timeliness (early/late prediction) and accuracy.
\underline{Third}, we further propose a novel hierarchical GNN that does not sacrifice timeliness but largely improves prediction accuracy, significantly outperforming HLS tools.
% Agile hardware development requires early and accurate circuit quality evaluation.
% Existing work makes performance evaluation/prediction after High-level Synthesis (HLS) requiring extensive feature engineering.
% To expedite circuit evaluation \textit{at the earliest stage}, we propose a rapid and accurate performance modeling, exploiting the representation power of graph neural networks (GNNs) by representing C/C++ programs as graphs.
% The contribution of this work is three-fold.
% \underline{First}, we build a standard benchmark containing 40k C synthesizable programs including three sets of real-world HLS benchmarks, each program implemented on FPGA with ground-truth performance metrics.
%\underline{Second}, we formulate the prediction problem on graphs and propose GNNs for modeling. We propose and discuss different prediction strategies, with trade-offs between prediction timeliness (early/late prediction) and accuracy.
% \underline{Third}, we further propose a novel hierarchical GNN that does not sacrifice timeliness but largely improves prediction accuracy, significantly outperforming HLS tools.
We apply extensive evaluations for both synthetic and \textit{unseen} real-case programs; our proposed predictor largely outperforms HLS by up to 40$\times$ and excels existing predictors by 2$\times$ to 5$\times$ in terms of resource usage and timing prediction.
\end{abstract}

%\begin{CCSXML}
%\end{CCSXML}

%\ccsdesc[500]{Computer systems organization~Embedded systems}

%\keywords{Graph Neural Network; High-Level Synthesis; Performance Prediction; Program Representation}

\maketitle

\vspace{-5pt}
\section{Introduction}

One essential requirement for agile hardware development is to evaluate circuit design quality quickly and accurately for rapid optimization iterations.
Traditional EDA tools usually take hours to days to accurately evaluate circuit quality with extensive manual efforts.
Although high-level synthesis (HLS) tools can greatly speed up circuit design, they still need minutes to hours for design synthesis, and can be largely inaccurate in terms of circuit quality evaluation~\cite{wu2021ironman}.
Given the strong need for hardware agile development and productivity boost, a quick and accurate performance evaluation at earliest stage, even before HLS, is highly expected.

\begin{figure}
    \centering
    \includegraphics[width=0.98\linewidth]{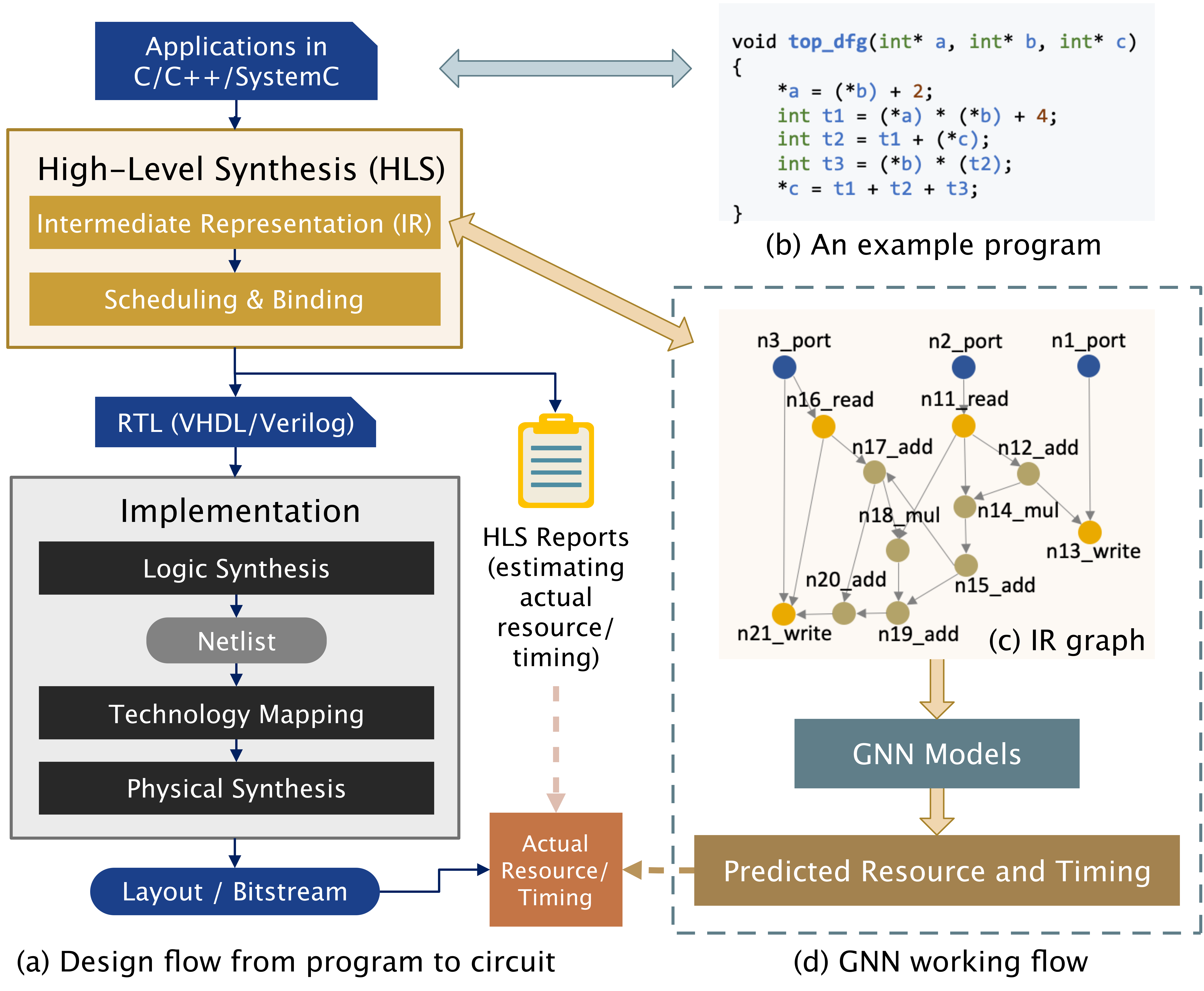}
    %\vspace{-10pt}
    \caption{The overall performance prediction flow.
    (a) Design flow starting from behavioral programs to hardware circuits. 
    (b) An example program written in C.
    (c) The intermediate representation (IR) graph extracted by compiler front-ends.
    (d) The working flow of GNNs, predicting \textit{actual} resource usage and timing merely based on raw IR graphs.}
    %\vspace{-20pt}
    \label{fig:overall}
\end{figure}

Prior work has investigated circuit performance evaluation before or after HLS, to predict synthesized or implemented design metrics such as resource usage, timing, power, and area.
Analytical models are classic approaches~\cite{zhao2017comba, zhao2019performance, perina2019lina} but they only work for highly regular dataflow such as perfect loops and arrays.
Recent ML approaches have become promising in estimating the actual design performance \cite{wu2021survey}.
Pyramid~\cite{makrani2019pyramid} assembled multiple ML models for resource and timing prediction.
%Common supervised learning techniques, e.g., linear regression, artificial neural networks (ANNs), gradient boosted regression trees, support vector machines, or random forest, can be used for routing congestion estimation in HLS \cite{zhao2019machine}, post-implementation resources on FPGAs \cite{dai2018fast}, 
%Zhao \textit{et al.}~\cite{zhao2019machine} propose to use linear regression, artificial neural networks, and gradient boosted regression trees for routing congestion estimation in HLS;
%similar ML techniques are also adopted by Dai \textit{et al.}~\cite{dai2018fast} to predict post-implementation resources on FPGAs.
%Pyramid~\cite{makrani2019pyramid} is another ML-based framework for resource and timing prediction with four ML models including linear regression, artificial neural network, support vector machine, random forest, and an ensemble model combining the four.
%HL-Pow~\cite{lin2020hl} is an ML-based predictor for power modeling of HLS.
Both HLSPredict~\cite{o2018hlspredict} and XPPE~\cite{makrani2019xppe} are ANN-based cross-platform performance predictors that estimate the HLS design performance on FPGAs.

Despite the great success, most of the ML-based methods rely on intensive and empirical feature engineering: a large number of features must be obtained from HLS synthesis report or the intermediate results of a partially executed implementation process, which is still time-consuming.
Therefore, in this work, we aim to approach HLS performance prediction \textit{at its earliest stage with least features} right after front-end compilation.
Since programs are usually represented as intermediate representation (IR) graphs at early stage, we exploit the representation power of graph neural networks (GNNs) and adopt various GNNs for timely performance prediction based on IR graphs.
Fig.~\ref{fig:overall} shows the overall prediction flow: we extract IR graphs right after HLS front-end compilation, and directly predict the actual circuit performance that are expected to be obtained after implementation.

To comprehensively investigate this problem,
we propose prediction algorithms at different stages of HLS and discuss their trade-offs of prediction accuracy and efficiency: the earlier the prediction is, the more beneficial for agile design but the less information.
We then propose a novel hierarchical GNN-based predictor with great trade-off, which can \textit{predict at earliest stage but still with sufficient domain-specific information}.
Further, to benefit follow-up researches, we standardize the problem formulation and develop a rich benchmark suite.
We summarize our contributions as follows:
%\vspace{-5pt}
\begin{itemize}[leftmargin=*]
    \item {\textbf{Benchmarking.} We build a standard benchmark containing 40k C programs, each with an IR graph. The programs are synthesized by HLS tool and implemented on FPGA implementation to get their actual resource and performance. Three sets of real-world benchmarks are included for generalization evaluation.
    }
    \item {\textbf{Modeling.} To study the trade-offs of prediction accuracy and efficiency, on IR graphs, we first propose two approaches using GNNs: (1) \textit{off-the-shelf approach} at earliest stage with least domain-specific information; (2) \textit{knowledge-rich approach} at later stage with HLS auxiliary information to improve prediction accuracy. We aim to provide domain insights for future GNN design.
    }
    \item {\textbf{Advancing.} We propose the third \textit{knowledge-infused approach}, a novel hierarchical GNN, that inherits both advantages of earliest prediction as well as domain knowledge.
    The model is composed of a node-level classification task and a graph-level regression task, which first classifies the resource type and then regresses the resource usage values.
    It largely improves the prediction accuracy with zero overhead at inference time.
    }
    \item {\textbf{Evaluation}.} We apply extensive evaluations for both synthetic and \textit{unseen} real-case programs; our proposed predictor largely outperforms HLS by up to 40$\times$ and excels existing predictor IronMan~\cite{wu2021ironman} by 2$\times$ to 5$\times$ in terms of resource usage and timing prediction.
\end{itemize}

%The benchmark suite and all the GNN models are publicly available at Github~\footnote{anonymous link}.

\begin{figure*}[tb]
     \centering
     %\vspace{-26pt}
     %\includegraphics[width=\textwidth]{figures/gnn.pdf}
     \includegraphics[width=\textwidth]{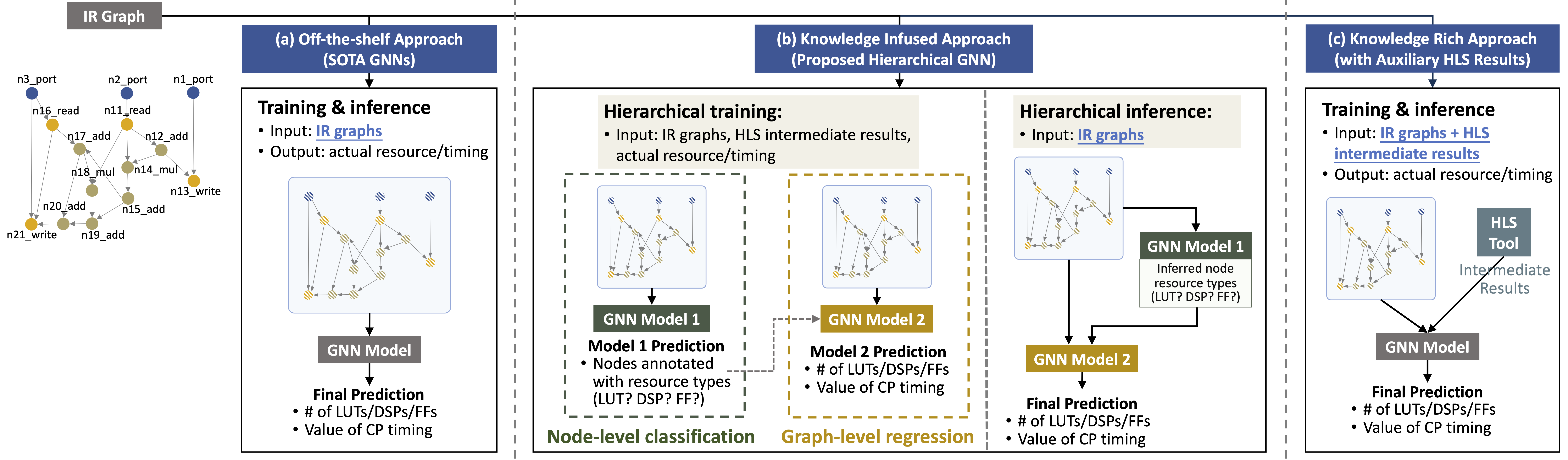}
     %\vspace{-15pt}
     \caption{Our three proposed approaches: (a) off-the-shelf approach at the earliest stage for prediction; (b) knowledge-infused approach also at the earliest stage but with self-inferred domain-specif information; (c) knowledge-rich approach. }
     \label{fig:gnn}
     %\vspace{-8pt}
\end{figure*}

%\vspace{-8pt}
\section{Performance Prediction Strategies}

There are two fundamental questions for performance prediction: \textbf{when} and \textbf{how}.

\noindent
\textbf{When to predict.} 
Performance prediction for a circuit design can be conducted at different stages in the synthesis flow, e.g., before or after HLS or during implementation. 
For instance, HLS tools take in behavioral language, convert to RTL, and produce a synthesis report, which is an early estimation of the final performance, as shown in Fig.~\ref{fig:overall}.
However, it is observed that HLS prediction (synthesis report) can be largely inaccurate~\cite{ustun2020accurate, wu2021ironman} even after minutes or hours of synthesis.
Both Pyramid~\cite{makrani2019pyramid} and XPPE~\cite{makrani2019xppe} make predictions for resource usage and timing after HLS by extracting features from HLS synthesis reports, while HLSPredict~\cite{o2018hlspredict} predicts FPGA cycle count and power before HLS.
Typically, an early and timely prediction is beneficial for agile development, but with less useful domain-specific knowledge.

\noindent
\textbf{How to predict.}
Existing ML-based prediction approaches attempted linear regression, artificial neural network, support vector machine, random forest, Lasso, and assembled models.
Although promising, these models require rich features as model inputs and thus heavy feature engineering is needed.
For instance, Pyramid/XPPE and HLSPredict requires up to 183 and 75 features, respectively, which can only be obtained by \textit{actually running HLS or CPU/FPGA sub-trace generation}.
Therefore, these strategies do not easily \textit{generalize} to new designs rapidly.

\noindent
\textbf{Our prediction strategy.}
%On the one hand, the earlier the prediction is, the more beneficial for shortening the evaluation-optimization iteration cycle;
%On the other hand, it is also more likely to be inaccurate, because a later prediction allows more domain/design-specific features to be extracted.
Our goal is to assist \textit{agile hardware development} by making the performance prediction \textit{as early as possible} and also \textit{as accurate as possible}.
Specifically, we focus on predicting the \textit{actual} design performance values on FPGA, including resource utilization and critical path timing.

To address the existing limitations (late prediction, hard to generalize) and to meet our goal, we propose three early prediction approaches, each with different amount of domain-specific knowledge leading to varied prediction accuracy.
%Especially, we introduce our best prediction strategy, that can predict \textit{at the earliest stage but with more domain-specific knowledge} that largely help with improving the prediction accuracy.
Fig.~\ref{fig:gnn} illustrates the three approaches. To make it \textit{timely}, we perform prediction based on the intermediate representation (IR) graph of a program, i.e., data flow graph (DFG) and control data flow graph (CDFG), which can be quickly extracted after the front-end compilation~\cite{aho2007compilers, wolf2012computers} within seconds.
To make it \textit{generalizable}, we propose to apply graph neural networks (GNNs) on DFGs/CDFGs, exploiting the inductiveness of GNNs to make predictions for completely unseen designs without retraining.
Specifically, our three approaches are:
%\vspace{-5pt}
\begin{itemize}[leftmargin=*]
    \item {\textit{Off-the-shelf approach}. The first approach makes prediction at the earliest stage by taking the IR graph (DFG/CDFG) as the GNN model input, and directly predicts the design performance metrics. The features can be obtained after HLS front-end compilation, resulting in fastest prediction since the compilation usually takes only a few seconds.
    }
    \item {\textit{Knowledge-rich approach}. The second approach aims to make more accurate prediction by taking auxiliary domain information from intermediate HLS results: the resource usage associated with each node. The features must be obtained during HLS execution, resulting in late but more more accurate prediction.
    }
    \item {\textit{Knowledge-infused approach}. The third approach is a novel \textit{hierarchical GNN} that
    possesses the advantages of both the first and second: it not only makes earliest prediction but also can infuse self-inferred domain-specific knowledge with almost zero overhead during inference.
    Specifically, it takes the IR graph as inputs and makes predictions in two steps: node-level classification and graph-level regression. %In the first step, it classifies the resource type associated to each node in the IR graph; in the second step, it regresses on the entire graph for resource usage and timing prediction.
    }
%\vspace{-15pt}
\end{itemize}

The approach details are introduced in Section~\ref{sec:models}.
Prior to that, we introduce our benchmark suite first in Section~\ref{sec:benchmark}.

\vspace{-8pt}
\section{Benchmarking}
\label{sec:benchmark}
%\vspace{-5pt}

The goal of benchmarking is to facilitate more ML related researches to promote rapid performance prediction by providing abundant synthesizable programs together with \textit{actual} performance values on FPGA, i.e., after implementation.
%including resource utilization and critical path timing.

%\vspace{-8pt}
\subsection{Benchmark Format}

\noindent
\textbf{Input.}
We let the inputs of a predictor to be the IR graph of a program, which can be quickly extracted after the front-end compilation~\cite{aho2007compilers, wolf2012computers}.
In HLS, DFGs and CDFGs are the most common IR graphs.
Specifically, DFGs are extracted from from \textit{basic blocks}, a straight-line code sequence with no branches in except to the entry and no branches out except at the exit~\cite{hennessy2011computer};
CDFGs are extracted from programs with loops. %\footnote{To distinguish loops in programs and loops in graphs, we use \textit{control loops} in the following context to represent loops in programs.}, jumps, and branches.
DFGs are directed acyclic graphs without any structural loops, while CDFGs contain additional nodes and edges/loops for control dependency.

\noindent
\textbf{Node/Edge Features}.
The three proposed approaches use different sets of node features, as listed in Table~\ref{table1}.
After HLS front-end compilation, there are seven features immediately available for each node to be used by the first off-the-shelf approach, such as node category, bitwidth, and opcode.
For the knowledge-infused and knowledge-rich approaches, we include the resource type and resource value, respectively, for each node as node features. 
Notably, for the knowledge-infused approach, the auxiliary node features (i.e., resource type) are only used during training but not inference.
Each edge has two features, the discrete edge type in integers, and a binary signal marking whether this edge is a back edge.

\noindent
\textbf{Tasks/labels.}
We provide two types of tasks, a \textit{node-level classification} task, and a \textit{graph-level regression} task, where the former is easier than the later.
%\vspace{-5pt}
\begin{itemize}[leftmargin=*]
    \item {For the \textit{node-level classification} task, we assign each node in the DFG/CDFG a label indicating the resource type(s) that the node will use in its final implementation. 
    We consider three resource categories: \texttt{DSP}, \texttt{LUT}, and \texttt{FF}.
    A node can be implemented by zero, one, or more resource types. 
For example, a \texttt{sdiv} node may use both \texttt{DSP} and \texttt{LUT}; a \texttt{partselect} node uses \texttt{FF} only; a node uses nothing if it is for control, e.g., indicating a branch entry (\texttt{br}).
%Based on the given DFG and the node information, an ML model should be able to predict what resource a certain node will use.
We organize the resource type prediction as three binary classification tasks.
If a node falls into none of the three, it is regarded as empty, i.e., not associated with any resource.}

\item{For the \textit{graph-level regression} task, we label the entire graph using its implemented performance metric values.
%The performance predictors are expected to predict common hardware quality metrics, include resource usage and circuit timing (which determines its maximum working frequency).
We consider four metrics for regression: \texttt{DSP}, \texttt{FF}, \texttt{LUT}, \texttt{CP}.
The first three are integer numbers indicating how much these resources are used; the last one is critical path timing slack in fractional number, determining the FPGA's maximum working frequency.}
%We expect that an ML model could predict these metrics based on the raw DFG/CDFG with minimum information.}
\end{itemize}
%\vspace{-5pt}

%\vspace{-5pt}
\subsection{Benchmark Generation.}
We construct the benchmark suite including synthetic and synthesizable C programs as well as real-world HLS applications.
The synthetic programs fall into two categories, basic blocks that derive DFGs, and programs with control loops and branches that derive CDFGs.
All of the synthetic programs are generated by a C program generator \texttt{ldrgen} \cite{barany2017liveness}. %, a plugin of Frama-C \cite{cuoq2012frama,kirchner2015frama}.
There are 19,120 and 18,570 C programs in the DFG and CDFG datasets, respectively, for graph-level tasks.
The node-level dataset contains more than 660k nodes derived from DFGs and CDFGs.
In addition, there are three sets of real-world HLS applications: MachSuite~\cite{reagen2014machsuite}, CHStone~\cite{hara2009proposal}, and PolyBench/C~\cite{PolyBench}, consisted of 16, 10, and 30 different applications, respectively.
The real-world applications are used for generalization evaluation of GNN models.

\begin{table}[tp]
\caption{Node features and example values.}
%\vspace{-12pt}
\label{table1}
\footnotesize
\centering
    \renewcommand{\arraystretch}{0.85}
    \setlength{\tabcolsep}{1.2pt}
\begin{tabular}{c | c | c }
\toprule
\textbf{Feature} & \textbf{Description} & \textbf{Values} \\ \midrule
\multicolumn{3}{c}{Off-the-shelf approach with minimum information} \\ \hline
Node type     &  General node type & \makecell[c]{\texttt{operation nodes}, \texttt{blocks},  \\ \texttt{ports}, \texttt{misc}}\\ 
\rowcolor[HTML]{EFEFEF}
Bitwidth          &  Bitwidth of the node  & \texttt{0}$\sim$\texttt{256}, \texttt{misc}\\
Opcode type   &  \makecell[c]{Opcode categories \\based on LLVM}  & \makecell[c]{\texttt{binary\_unary}, \texttt{bitwise}, \\ \texttt{memory}, etc.}  \\ 
\rowcolor[HTML]{EFEFEF}
Opcode            &  Opcode of the node &  \texttt{load}, \texttt{add}, \texttt{mux}, \texttt{xor}, \texttt{icmp}, etc. \\
Is start of path  & \makecell[c]{Whether the node is \\ the starting node of a path}  & \texttt{0}, \texttt{1}, \texttt{misc}   \\
\rowcolor[HTML]{EFEFEF}

Cluster group  & Cluster number of the node & \texttt{-1}$\sim$\texttt{256}, \texttt{misc}\\

\midrule
\multicolumn{3}{c}{Knowledge-infused and knowledge-rich approach} \\ \hline

DSP  & DSP used for this node? & binary/integer values, \texttt{misc}\\
\rowcolor[HTML]{EFEFEF}
LUT  & LUT used for this node? & binary/integer values, \texttt{misc}\\
FF   & FF used for this node? & binary/integer values, \texttt{misc}\\
\bottomrule
\end{tabular}
%\vspace{-15pt}
\end{table}

%\vspace{-5pt}
\section{Modeling and Advancing with GNNs}
\label{sec:models}
%\vspace{-3pt}
In this section, we introduce the three proposed GNN-based approaches with trade-offs between timeliness and accuracy.
GNNs operate by propagating information along the edges of a given graph.
By stacking multiple GNN layers, each node can receive information from multi-hop neighbors and locally characterize the corresponding receptive field for node-level tasks.
Graph pooling then summarizes global information to perform graph-level prediction tasks.

%\vspace{-8pt}
\subsection{Modeling: Off-the-Shelf Approach with State-of-the-Art GNN Models}
%\vspace{-5pt}

In the off-the-shelf approach, we screen several state-of-the-art GNN models, aiming to identify (1) which properties of existing GNN models would help with resource/timing prediction and (2) how domain-specific insights can be combined with these properties to improve prediction accuracy.
14 different GNN models are selected from four categories based on how topological and relational information in graphs are exploited, which are briefly introduced as follows.
%\vspace{-3pt}
\begin{itemize}[leftmargin=*]
    \item \textbf{Graph convolutional network (GCN) and variants}: (1) GCN \cite{kipf2016semi}; (2) GCN equipped with a virtual node \cite{gilmer2017neural}; (3) SGC~\cite{wu2019simplifying}, a simplified version of GCN;
    (4) GraphSage \cite{hamilton2017inductive}, a variant of GCN sampling a fixed number of neighboring nodes to keep the computational footprint consistent;
    (5) ARMA~\cite{bianchi2021graph}, a variant of GCN with auto-regressive moving average filters;
    (6) PAN~\cite{ma2020path}, a generalization of GCN assigning trainable weights to each path based on its length. 
    Previous work \textsc{IronMan}~\cite{wu2021ironman} also has a similar GCN-based performance predictor.
    \item \textbf{Graph isomorphism network (GIN) and variants.} 
    (1) GIN \cite{xu2018powerful}, provably as powerful as Weisfeiler-Lehman graph isomorphism test;
    (2) GIN with a virtual node \cite{gilmer2017neural};
    (3) PNA~\cite{corso2020principal}, leveraging complementary aggregators to better understand graph structures and retain neighborhood information, especially for a continuous input feature space.
    \item \textbf{Employing multi-relational information.} 
    (1) GAT~\cite{velivckovic2017graph}, using attention mechanisms to implicitly assign different importance to nodes in the same neighborhood;
    (2) GGNN~\cite{li2015gated}, using trainable edge-dependent weights with gated recurrent units;
    (3) RGCN~\cite{schlichtkrull2018modeling}, using edge-dependent weight with non-linearity activation.
    \item \textbf{Inspired from vision tasks.}
    (1) Graph U-Net \cite{gao2019graph}, using an encoder-decoder structure on graphs;
    (2) GNN-FiLM~\cite{brockschmidt2020gnn}, combining feature-wise linear modulation (FiLM) with the message passing procedure.
\end{itemize}
%\vspace{-3pt}

To fairly evaluate these models, we use the same GNN structure (e.g., embedding, layer count) but with different types of GNN layers.
The goal is to directly predict actual resource/timing based on IR graphs without invoking HLS.
This approach makes earliest prediction since the HLS front-end compilation is the very first step of an EDA design flow.
While with the best timeliness, the accuracy is compromised due to the ignorance of device-specific information.

%\vspace{-8pt}
\subsection{Modeling: Knowledge Rich Approach with Selected GNN Models}
%\vspace{-5pt}

To include device information revealed during the design flow, we devise the knowledge rich approach, which takes both IR graphs and auxiliary information from intermediate HLS results as inputs, as shown in Fig.~\ref{fig:gnn}(c).
The auxiliary information from HLS tools indicates both the type(s) of resource and the exact number of each resource used in final implementation for every node in IR graphs.
As each node is marked with pre-characterized resource estimations, GNN models pay more attention to resource interference/sharing among nodes, achieving much better prediction accuracy.

Armed with rich domain knowledge, this approach emphasizes more on prediction accuracy, especially for resource estimation, yet compromises timeliness since HLS tools do take some time to generate intermediate results.

%\vspace{-10pt}
\subsection{Advancing: Knowledge Infused Approach with Hierarchical GNN Models}
%\vspace{-5pt}

To strike a balance between timeliness and accuracy, we propose the knowledge infused approach with hierarchical GNN models.
As depicted in Fig.~\ref{fig:gnn}(b), the resource/timing prediction task is disentangled into two sub-tasks: node-level classification that annotates resource types associated with each node, and graph-level regression that predicts actual resource/timing with the annotated graphs.
During the hierarchical training, the node-level classification takes IR graphs as inputs, and the domain knowledge is infused by providing labels to each node that denote resource types used in final implementation based on HLS intermediate results; 
the graph-level regression then takes both IR graphs and ground-truth resource types as inputs, aiming to convey the infused domain knowledge from node-level to graph-level tasks and to improve final prediction accuracy.
During the hierarchical inference, the only required inputs are IR graphs:
first, the node-level GNN model infers resource types for each node;
second, combining the node-level inference results with original IR graphs, the graph-level regression grasps self-inferred domain knowledge to perform final predictions.

Taking advantages of knowledge infusion during training, this approach demonstrates a great balance between timeliness and accuracy: predicting resource/timing from the earliest stage and simultaneously
adopting adequate domain information to improve prediction accuracy.

%\vspace{-10pt}
\section{Experiment}
%\vspace{-5pt}
\subsection{Experimental Setup}
%\vspace{-5pt}
\label{sec:exp_setup}
All GNN models are implemented with Pytorch Geometric \cite{Fey/Lenssen/2019}.
The ground-truth (actual) resource usage (LUT/DSP/FF) and CP timing are synthesized by Vitis HLS \cite{VitisHLS} and implemented by Vitis \cite{Vitis}.
DFG and CDFG datasets are randomly split into 80\% train, 10\% validation and 10\% test; real-world benchmarks are only used for generalization evaluation.
Each GNN is empirically set as five layers with a hidden-dimension size of 300.
For graph-level regression, sum or mean pooling is used to derive graph representations, followed by a feed-forward network with the structure 300-600-300-1.
Models are trained using Adam optimizer for 100 epochs.
Learning rates, dropout and other hyper-parameters are tuned on the validation set. 
Each model is trained with five runs using different random number seeds and we report the average of three with least validation error.

%\vspace{-8pt}
\subsection{Modeling: SOTA GNN Analysis}
%\vspace{-5pt}

We launch discussions of the off-the-shelf approach from three aspects: how different applications (i.e., graphs) influence prediction accuracy, which properties of existing GNN models would help improve accuracy, and what domain-specific insights can be derived to facilitate future graph representation learning on fast evaluation in EDA tasks.

\noindent
\textbf{Different graphs: DFG vs. CDFG}.
Table \ref{table:result} exhibits mean absolute percentage error (MAPE) of predictions on DFGs and CDFGs from synthetic programs. 
The MAPE on CDFGs is larger than that on DFGs, which attributes to two major reasons.
First, DFGs have no loops but CDFGs typically include a considerable number of loops.
Since message-passing-based GNN models have limited expressiveness and are not better than the 1-Weisfeiler-Lehman isomorphism test \cite{maron2019provably}, they are not excelled to handle graphs with many loops.
Second, control signals introduce additional nodes/edges that represent control states and dependency, which are seemingly unrelated to resource usage. These nodes/edges easily confuse GNN models during resource prediction.

\noindent
\textbf{Model analysis}.
PNA and RGCN generally show superior performance, implying two takeaways.
First, the relational information (i.e., edge information) is important in IR graphs, since they represent data or control dependency, or a mix of both, which is a critical basis in logic synthesis and impacts resource allocation.
Second, equipped with multiple aggregators, PNA is more powerful to characterize different neighborhood information, thus making better predictions.

\noindent
\textbf{Domain-specific insights}.
\underline{\textit{(1) Resource.}}
Among three types of resource, DSPs are mainly used for computation; 
FFs often relate to memory operations and small arrays;
LUTs may appear in computation, memory or control nodes.
The key to making precise DSP prediction is to distinguish major computation nodes that are most likely to use DSPs. 
For instance, a multiplication node with a large bitwidth tends to use DSPs, while divisions and bitwise operations prefer LUTs.
Similarly, effective extraction of memory-related nodes would greatly benefit FF predictions.
Since LUTs are involved in the entire graph (as computation units and glue logic to circuit components), graph-level understanding is important.
To briefly summarize, it is helpful to carefully characterize neighborhood information from each node's predecessors, successors, itself, and their relations, such that the sophisticated mapping rules from heterogeneous nodes to resource usage can be clearly understood and quantitatively learned.
\underline{\textit{(2) Timing.}}
Compared with resource predictions, CP timing predictions show relatively lower MAPE and better consistency between DFGs and CDFGs. 
A probable reason is that CP timing is local information and thus is insensitive to graph sizes as long as the critical path segment can be recognized.
%the CP timing can be accurately predicted.

\begin{table}[t]
%\vspace{-5pt}
\caption{MAPE of graph-level regression with different GNN models on DFG and CDFG datasets. The top two performant models are marked in bold.}
%\vspace{-10pt}
\label{table:result}
\centering
    \footnotesize
    \renewcommand{\arraystretch}{0.85}
    \setlength{\tabcolsep}{1.8pt}
\begin{tabular}{c|cccc|cccc}
\toprule
                   & \multicolumn{4}{c|}{\textbf{DFG}}   & \multicolumn{4}{c}{\textbf{CDFG}}       \\ 
                   & \textbf{DSP}      & \textbf{LUT}      & \textbf{FF}      & \textbf{CP} 
                   & \textbf{DSP }     & \textbf{LUT}      & \textbf{FF}      & \textbf{CP}       \\ \midrule
\textbf{GCN}                & 16.31\%  & 16.49\%  & 21.27\%   & \textbf{6.12\%} 
                   & 25.30\%  & 28.64\%  & 38.34\%   & 8.79\%          \\
\rowcolor[HTML]{EFEFEF}
\textbf{GCN-V}           & 15.72\%  & 15.93\%  & 21.64\%   & 6.36\%          
                   & 17.31\%  & 33.93\%  & 39.94\%   & \textbf{8.13\%} \\
\textbf{SGC}                & 42.12\%  & 23.93\%  & 30.61\%   & 7.92\%          
                   & 44.01\%  & 60.87\%  & 53.50\%   & 10.32\%         \\
\rowcolor[HTML]{EFEFEF}
\textbf{SAGE}          & 15.18\%  & 14.01\%  & 17.11\%   & \textbf{6.12\%} 
                   & 17.01\%  & 28.09\%  & 39.11\%   & \textbf{8.25\%} \\
\textbf{ARMA}               & 19.12\%  & 13.46\%  & 16.87\%   & 6.50\%          
                   & 18.47\%  & \textbf{25.21\%} & 32.15\%  & 8.42\%    \\
\rowcolor[HTML]{EFEFEF}
\textbf{PAN}                & 15.24\%  & 14.13\%  & 17.23\%   & 6.38\%          
                   & 16.88\%  & 32.65\%  & 44.36\%   & 8.54\%          \\
\textbf{GIN}                & 15.52\%  & 16.10\%  & 22.08\%   & 6.58\%          
                   & 15.47\%  & 28.48\%  & 38.82\%   & 8.76\%          \\
\rowcolor[HTML]{EFEFEF}
\textbf{GIN-V}           & 15.04\%  & 16.17\%  & 23.09\%   & 6.40\%          
                   & 17.94\%  & 29.40\%  & 48.64\%   & 8.59\%          \\
\textbf{PNA}  & \textbf{12.65\%} & \textbf{11.64\%} & \textbf{14.41\%}  & 6.26\% 
     & \textbf{14.71\%} & \textbf{22.86\%} & \textbf{26.47\%}  & 8.87\% \\
\rowcolor[HTML]{EFEFEF}
\textbf{GAT}                & 26.22\%  & 22.64\%  & 27.74\%   & 8.30\%  
                   & 28.66\%  & 46.19\%  & 54.73\%   & 10.32\%         \\
\textbf{GGNN}               & 15.40\%  & 13.64\%  & 16.94\%   & 6.47\%          
                   & 16.28\%  & 28.05\%  & 31.88\%   & 8.50\%          \\
\rowcolor[HTML]{EFEFEF}
\textbf{RGCN} & \textbf{13.27\%} & 13.03\% & \textbf{15.09\%} & 6.14\% 
     & \textbf{15.03\%} & 26.33\% & \textbf{25.52\%} & 8.72\% \\
\textbf{UNet}               & 18.40\%  & 14.90\%  & 19.17\%   & 6.61\% 
                   & 18.92\%  & 32.83\%  & 53.06\%  & 9.02\%          \\
\rowcolor[HTML]{EFEFEF}
\textbf{FiLM }              & 20.05\%  & \textbf{12.50\%} & 16.94\%  & 6.27\% 
                   & 17.42\%  & 26.97\%  & 27.35\%   & 8.67\%         \\
\bottomrule
\end{tabular}
%\vspace{-20pt}
\end{table}

\subsection{Advancing: Comparison of Three Approaches}
%\vspace{-5pt}
We first discuss the results of the knowledge infused approach, and then comprehensively compare the three proposed approaches with HLS and prior work IronMan~\cite{wu2021ironman}.

\noindent
\textbf{Knowledge infused approach.}
Essentially, using GNNs to predict actual resource/timing from IR graphs is to approximate the set of sophisticated heuristics and mapping rules used by HLS scheduling/binding and logic/physical synthesis during design flow.
The evaluation of the off-the-shelf approaches indicate that \textit{plug-in application of GNNs cannot well approximate such underlying rules}.
Thus, in addition to infusing domain knowledge during training, another motivation of the hierarchical structure in the knowledge infused approach is to \textit{divide and conquer}.
The complicated performance prediction task is decoupled as two simpler sub-tasks: 
for \textit{node-level classification}, Table \ref{table:node_result} shows prediction accuracy of classifying resource types, where high accuracy is achieved for most of the cases since local neighborhood characterization is enough for node-level resource type classification;
for \textit{graph-level regression}, Table \ref{table:hierarchical} displays MAPE of predictions on synthetic programs, showing an obvious accuracy boost compared with the off-the-shelf approach.

With the hierarchical training, both the node-level and the graph-level GNN models in Fig.~\ref{fig:gnn}(b) are approximating simplified design heuristics.
Specifically, the node-level classification aims to understand the preference of resource types on different nodes; the graph-level regression focuses on globally estimating resource sharing and interference among nodes.
With the hierarchical inference, the domain knowledge infused during training can be self-inferred when encountering unseen designs, leading to improved prediction accuracy from the earliest design stage.

\noindent
\textbf{Accurate, timely, and generalizable.}
The three approaches explore different trade-offs between timeliness and accuracy.
Intuitively, the more domain information is leveraged, the more accurate predictions are provided, whereas the longer time would be taken for feature collection.
The off-the-shelf approach makes predictions from the earliest stage simply with IR graphs, at the cost of accuracy loss due to ignorance of domain knowledge.
The knowledge rich approach provides the best prediction accuracy, yet has to wait for HLS tools providing intermediate results, sacrificing timeliness.
The knowledge infused approach shows a balance: infusing adequate domain knowledge during training, and making predictions from the earliest stage during inference.

Generelization capability is a key indicator of whether an ML/GNN-based approach can be widely applied for certain EDA tasks.
Table \ref{table:real} shows MAPE of the three proposed approaches and Vitis HLS on real-case applications.
Compared with Vitis HLS, our approaches significantly improve prediction accuracy especially for LUT/FF usage and CP timing. Specifically, PNA-based knowledge-infused approach outperforms HLS by 1.2$\times$ to 40.6$\times$, while PNA-based knowledge-rich approach outperforms HLS by 1.7$\times$ to 51.4$\times$.
Note that since IronMan~\cite{wu2021ironman} is a variant of off-the-shelf GCN, whose performance is inferior to RGCN, the results in Table~\ref{table:real} imply that our proposed hierarchical GNN outperforms IronMan by at least 2.1$\times$ to 5.0$\times$.

Such results empirically demonstrate (1) generalization capability not only from seen to unseen designs but also from synthetic to realistic applications, (2) accuracy and timeliness conspicuously surpassing HLS tools.

% Such results empirically demonstrate that given the current message passing mechanisms and model designs, \textit{GNNs are capable to learn a simplified version of the sophisticated heuristics and mapping rules used in scheduling/binding and logic/physical synthesis, at least much better than HLS tools.}

% potential challenges

%\vspace{-5pt}
\section{Conclusion}
%\vspace{-5pt}

In this work, we discussed three approaches for early circuit performance prediction using GNNs:
(1) the off-the-shelf approach, making earliest prediction with least domain-specific information, showing 
on-par performance with HLS;
(2) the knowledge-rich approach, making late prediction after HLS with auxiliary information, showing significantly better performance than HLS;
(3) the knowledge-infuse approach, making earliest prediction in a two-step hierarchical manner with self-inferred knowledge, still significantly outperforming HLS.
We also constructed a standard benchmark suite for facilitating future researches.
This work not only demonstrated the great potential of GNN in EDA, but also advanced the GNN design by proposing innovative architectures.

\begin{table}[tb]
%\vspace{-20pt}
\caption{Prediction accuracy of node-level resource classification with four different GNN models on DFGs, CDFGs and real-case applications.}
\label{table:node_result}
\centering
    \footnotesize
    \renewcommand{\arraystretch}{0.87}
    \setlength{\tabcolsep}{0.3pt}
%\vspace{-10pt}
\begin{tabular}{c|ccc|ccc|ccc}
\toprule
 & \multicolumn{3}{c|}{\textbf{DFG}} & \multicolumn{3}{c|}{\textbf{CDFG}} & \multicolumn{3}{c}{\textbf{Real Case}} \\
 & \textbf{DSP} & \textbf{LUT} & \textbf{FF} & \textbf{DSP} & \textbf{LUT} & \textbf{FF} & \textbf{DSP} & \textbf{LUT} & \textbf{FF} \\ \midrule
\textbf{GCN} & 93.79\% & 84.84\% & 88.66\% & 83.00\% & 77.01\% & 64.74\% & 79.70\% & 81.83\% & 86.82\% \\
\rowcolor[HTML]{EFEFEF}
\textbf{SAGE} & 93.06\% & \textbf{87.32\%} &\textbf{ 92.09\%} & 85.65\% & 78.41\% & 60.40\% & 87.39\% & 86.44\% & 55.88\% \\
\textbf{GIN} & 93.80\% & 84.93\% & 91.57\% & 79.24\% & 73.05\% & 65.78\% & 74.70\% & 75.53\% & 72.24\% \\
\rowcolor[HTML]{EFEFEF}
\textbf{RGCN} & \textbf{93.91\%} & 87.13\% & 91.52\% & \textbf{85.80\%} & \textbf{78.46\%} & \textbf{68.92\%} & \textbf{90.82\%} & \textbf{88.83\%} & \textbf{91.55\%} \\ \bottomrule
\end{tabular}
%\vspace{-15pt}
\end{table}

\begin{table}[tb]
%\vspace{-20pt}
\caption{MAPE of the three proposed approaches with RGCN/PNA on DFG and CDFG datasets. The default notation means the off-the-shelf approach; -I means the knowledge infused approach; -R means the knowledge rich approach.}
\label{table:hierarchical}
\centering
    \footnotesize
    \renewcommand{\arraystretch}{0.85}
    \setlength{\tabcolsep}{1.9pt}
\vspace{-10pt}
\begin{tabular}{c|cccc|cccc}
\toprule
 & \multicolumn{4}{c|}{\textbf{DFG}} & \multicolumn{4}{c}{\textbf{CDFG}} \\
 & \textbf{DSP} & \textbf{LUT} & \textbf{FF} & \textbf{CP} & \textbf{DSP} & \textbf{LUT} & \textbf{FF} & \textbf{CP} \\ \midrule
\textbf{RGCN}\tablefootnote{An advanced version of \textsc{IronMan} by considering relational information.}
                 & 13.27\% & 13.03\% & 15.09\% & 6.14\% 
                 & 15.03\% & 26.33\% & 25.52\% & 8.72\% \\
\rowcolor[HTML]{DAE8FC} 
\textbf{RGCN-I}  & 10.60\% & 10.25\% & 12.47\% & 5.70\% 
                 & 12.65\% & 20.55\% & 19.01\% & 6.78\% \\
\textbf{RGCN-R}  & 8.86\%  & 8.58\%  & 10.18\% & \textbf{4.91\%} 
                 & 10.98\% & 14.06\% & 16.65\% & \textbf{5.46\%} \\ \hline
\textbf{PNA}     & 12.65\% & 11.64\% & 14.41\% & 6.26\% 
                 & 14.71\% & 22.86\% & 26.47\% & 8.87\% \\
\rowcolor[HTML]{DAE8FC} 
\textbf{PNA-I}   & 8.26\%  & 5.10\%  & 7.58\%  & 5.51\% 
                 & 10.39\% & 14.12\% & 16.42\% & 6.54\% \\
\textbf{PNA-R}   & \textbf{7.06\%} & \textbf{4.02\%}  & \textbf{5.78\%}  & 5.39\% 
                 & \textbf{8.95\%} & \textbf{10.27\%} & \textbf{11.22\%} & 5.81\% \\ 
\bottomrule
\end{tabular}
%\vspace{-20pt}
\end{table}

\begin{table}[t]
%\vspace{-10pt}
\caption{Testing MAPE of the three proposed approaches with RGCN/PNA on real-case applications. }
\vspace{-10pt}
\label{table:real}
\centering
    \footnotesize
    \renewcommand{\arraystretch}{0.85}
    \setlength{\tabcolsep}{2.5pt}
\begin{tabular}{c|
>{\columncolor[HTML]{EFEFEF}}c |c
>{\columncolor[HTML]{DAE8FC}}c c|c
>{\columncolor[HTML]{DAE8FC}}c c}
\toprule
 & \textbf{HLS} & \textbf{RGCN} & \textbf{RGCN-I} & \textbf{RGCN-R} & \textbf{PNA} & \textbf{PNA-I} & \textbf{PNA-R} \\ \midrule
\textbf{DSP} & 26.07\%  & 45.61\%  & 40.89\% & 32.90\% & 40.06\% & 21.95\% & \textbf{15.20\%} \\
\textbf{LUT} & 871.56\% & 66.23\%  & 30.91\% & 24.08\% & 56.34\% & 21.45\% & \textbf{16.96\%} \\
\textbf{FF}  & 322.86\% & 101.20\% & 38.75\% & 27.72\% & 47.65\% & 20.10\% & \textbf{17.42\%} \\
\textbf{CP}  & 32.09\%  & 8.13\%   & 5.35\%  & 5.83\%  & 8.68\%  & 4.80\%  & \textbf{3.97\%} \\ \bottomrule
\end{tabular}
%\vspace{-10pt}
\end{table}
%\vspace{-5pt}
\bibliographystyle{ACM-Reference-Format}
\bibliography{ref}

\end{document}